\documentclass{article}
\usepackage{spconf,amsmath,graphicx}
\usepackage{subfig}

\title{CONCATENATED FEATURE PYRAMID NETWORK FOR INSTANCE SEGMENTATION}
\name{Yongqing Sun $^{\star}$ \qquad Pranav Shenoy K P $^{\star \dagger}$ \qquad Jun Shimamura $^{\star}$ \qquad Atsushi Sagata$^{\star}$}
\address{$^{\star}$ Media Intelligence Lab, NTT Corporation, Japan \\ $^{\dagger}$ Georgia Institute of Technology, USA
    \\ \footnotesize{\qquad\texttt{\{yongqing.sun.fb, jun.shimamura.ec, atsushi.sagata.hw\}@hco.ntt.co.jp} \qquad \texttt{pskp3@gatech.edu}}}

\begin{document}
%
\maketitle
\begin{abstract}


Low level features like edges and textures play an important role in 
accurately localizing instances in neural networks.
In this paper, we propose an architecture which improves feature pyramid networks
commonly used instance segmentation networks by incorporating low level features 
in all layers of the pyramid in an optimal and efficient way. Specifically, we 
introduce a new layer which learns new correlations from feature maps of 
multiple feature pyramid levels holistically and enhances the semantic
information of the feature pyramid to improve accuracy.
Our architecture is simple to implement in instance segmentation or object 
detection frameworks to boost accuracy. Using this method in Mask RCNN, our 
model achieves consistent improvement in precision on COCO Dataset with the
computational overhead compared to the original feature pyramid network.

\end{abstract}
\section{Introduction}
\label{sec:intro}

Instance segmentation is one of the most important developments in computer
vision. It combines object detection and semantic segmentation and finds its 
application in a wide variety of applications ranging from autonomous driving
to medical imaging to video surveillance. One of challenges faced by instance
segmentation is detecting and segmenting objects at vastly different scales.
An efficient way to overcome this challenge would be to create feature pyramids
from multiple layers of the CNN \cite{C1,C2,C16,C17}. This
type of framework combines low resolution but semantically strong features
with high resolution but semantically weak features in a top-down pathway with
lateral connections from lower layers. 

Mask RCNN\cite{C3} and Path Aggregation Networks or PANet\cite{C5} are popular 
state-of-the-art frameworks used for instance segmentation\cite{C15}. Mask RCNN 
extends Faster RCNN\cite{C4} by adding a Fully Cconvolutional Network\cite{C19} 
branch for predicting an object mask in parallel with 
the existing branch for bounding box recognition and utilizes feature pyramids\cite{C1}
to achieve high accuracy. PANet enhances this architecture by adding a 
bottom-up pathway with lateral connections after the top-down pathway along 
with other improvements to Mask RCNN. By adding a bottom-up pathway, the
features in low levels which are helpful for identifying large objects take
a shorter path to reach higher levels and improve localization. However, in both
of these frameworks, the features are added to subsequent layers one after the 
other and by using element-wise addition. Also due to this process of addition, there
are no layers which contain the correlation between high-level and low-level features.
Experiments have shown that utilizing correlations between different levels of
features can potentially further improve the performance of the network\cite{C24}.
Experiments and papers such as \cite{C22,C21}
have shown that concatenation is more flexible compared to element-wise addition 
and can improve the performance of the network. DenseNet\cite{C21} uses concatenations
or dense connections to achieve parameter efficiency and feature reuse which can 
give better performance with lesser or similar computational resources.

The motivation behind this paper is to improve the performance and mask quality
of the network by overcoming the drawbacks of existing frameworks. To achieve this,
we propose the following:
\begin{enumerate}
    \item A new convolutional layer to learn correlations between different levels
        of features.
    \item A bottom-up pathway to infuse low-level features from the lower 
        pyramid levels to the higher levels in a computationally efficient way.
\end{enumerate}

\begin{figure*}[htb]
\begin{minipage}[b]{1.0\linewidth}
  \centering
  \centerline{\includegraphics[width=15cm]{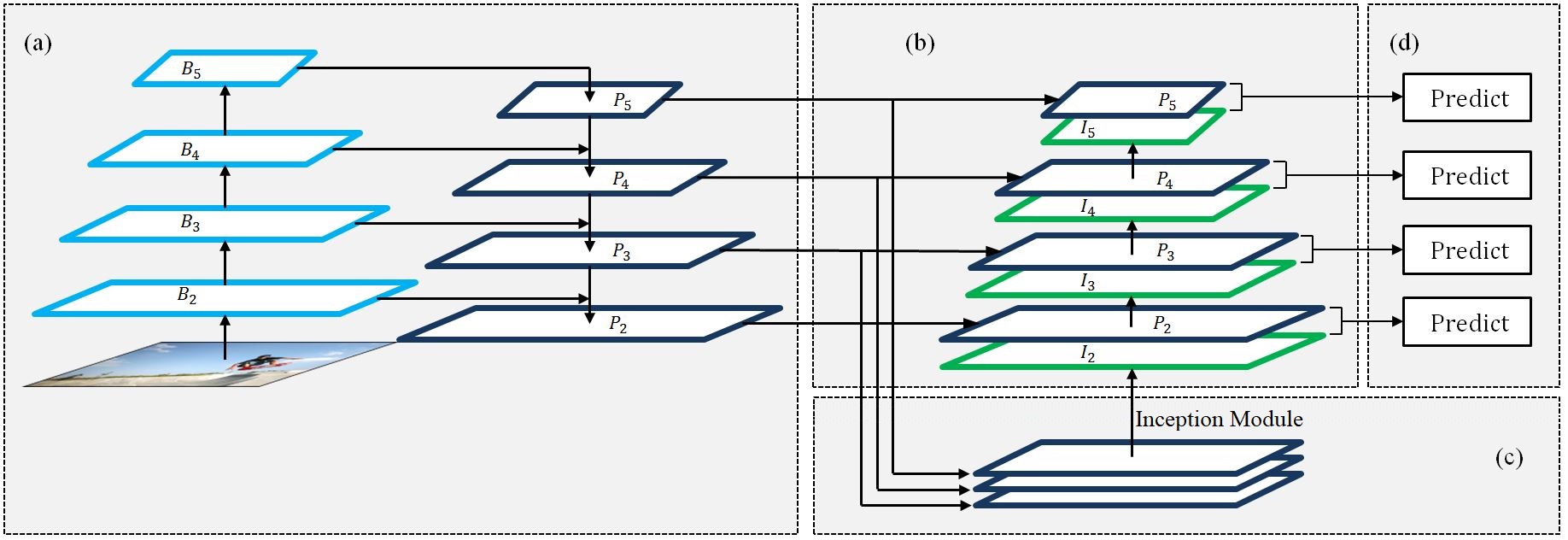}}
\end{minipage}
\centering
\caption{Illustration of our network. (a) FPN backbone. 
  (b) Bottom-up pathway. (c) Inception module. (d) ROI Align.}
\label{fig:res}
\end{figure*}


\subsection{Feature Pyramids}
\label{ssec:subhead}

Analysis by M. D. Zeiler and R. Fergus\cite{C7} on feature maps have shown that neurons in
the higher layers of the network are activated by entire objects or large regions of
objects while neurons in lower layers are more likely to be activated by edges, local
texture, patterns and other lower level features. The localization accuracy
of a framework can be further enhanced by propagating strong activations of
low-level features to higher layers since strong activations to edges or object
parts are good indicators to accurately localize objects; particularly small
objects. Hence by adding low-level features to higher levels of the feature 
pyramid, we can achieve better performance for mask generation.

Networks like FPN\cite{C1}, U-Net\cite{C20} and TDM\cite{C2} improve the accuracy 
by infusing features from lower layers. FPN augments a top-down path
with lateral connections creating a feature pyramid for building high-level
semantic feature maps at all scales. The top-down pathway hallucinates higher
resolution features by upsampling feature maps from higher pyramid levels and
adding them features from lower levels with lateral connections. Through this,
the features from higher features reach the lower layers of the pyramid.
However, lower level features do not reach the upper levels since the layers are 
added in a single direction. In figure 1, section (a) is the framework of the 
original feature pyramid network.

\subsection{DenseNet}
\label{ssec:subhead}

Densely connected convolutional \cite{C21} networks or DenseNet connects all layers in the 
network directly with each other using concatenation to ensure improved flow of 
information and gradients between layers. Each layer in the network obtains 
additional inputs from all preceding layers and passes on its own feature-map to
all subsequent layers. This network has direct access to the gradients from loss 
function and the original input signal leading to an implicit deep supervision\cite{C22}.
These dense connections also condense the model and make it easy to train and highly
parameter-efficient. Concatenating feature-maps learned by different layers increases
variation in input of subsequent layers and improves efficiency. Since the bottom
layers have a shorter path to the top layers, the gradients reach the bottom
layers more efficiently and reduce training error.



\section{Framework}
\label{sec:format}

\subsection{Concatenated Feature Pyramid Network}
\label{ssec:subhead}

Concatenated Feature Pyramid Network(CFPN) is an enhanced version of 
Feature Pyramid Network designed to overcome the drawbacks of FPN by 
adding an addition feature pyramid. 
In this additional pyramid, the features are added in the reverse direction, i.e., 
from bottom to top. To reduce computation and to further enhance the performance,
we have used a combination of concatenation and downsampling to propagate features.
The advantage of using concatenation over addition (which is used in 
bottom-up path augmentation \cite{C15}) is that the features are added more flexibly,
i.e., the network learns the optimal ratio to infuse the features which boosts
performance. However, we use element-wise addition in the original top-down pyramid,
since using concatenation did not impact the performance and also required more
computation. Finally, we append a 3x3 post-hoc convolution on each concatenated map 
([{I$_{i}$, P$_{i}$}]) to generate the final feature map. This is done to reduce the 
aliasing effect of upsampling in top-down layers.

To solve the issue of finding correlation between high-level and low-level features,
we introduce a convolutional layer between the top-down pyramid and the new bottom-up
pyramid structure as shown in figure 1(c). Here we upsample the top two layers of the top-down
pyramid and concatenate them with the third layer. We chose
not to include the bottom-most layer because this layer is concatenated and processed
just after the Inception module. Adding this layer increases cost and also
did not affect the final performance. Instead of using a 3x3 
convolutional layer, we chose to use an Inception module \cite{C8}. The Inception module 
\cite{C6} learns from cross-channel correlations and spatial correlations of the
feature map by using multiple kernel sizes for learning features with different
field of views\cite{C7}. This is particularly useful since the feature of the concatenated
layer contain features of different spacial dimensions due to upsampling and also 
due to their hierarchy in the backbone ResNet. The output of the Inception module is
then concatenated to the bottom-most layer of the top-down module as shown in
figure 1(b).

We take ResNet backbone as the basic structure and use {P$_{2}$, P$_{3}$, 
P$_{4}$, P$_{5}$} to denote the layers of the top-down pathway generated by the
FPN. The top most layer of the top-down pyramid can be represented as
\begin{equation}
P_{5} = H_{5}(B_{5})
\end{equation}
Where H$_{5}$ denotes the 1x1 convolution function followed by Relu activation.
The other lower layers are added with the higher layers using addition. Hence
these layers can be represented as
\begin{equation}
P_{l} = H_{l}(B_{l}) + U_{l}(P_{l-1})
\end{equation}
Where U$_{l}$ is the nearest neighbor upsampling function.
We use {I$_{2}$, I$_{3}$, I$_{4}$, I$_{5}$} to denote the layers of the 
newly generated Inception pyramid from the output of the Inception module. A
detailed illustration of the bottom-up pathway is given in figure 3.
The output of the Inception module forms the bottom-most layer of the 
Inception pyramid. It can be written as a function of concatenation
of P$_{5}$ to P$_{3}$.
\begin{equation}
I_{2} = G_{2}([P_{5}, P_{4}, P_{3}])
\end{equation}
Here \([.,.]\) is the concatenation operation and
G$_{2}$ denotes the function of the Inception module followed by Relu activation.
Each layer in the Inception pyramid I$_{i+1}$ is created by combining
lower layers of the Inception pyramid I$_{i}$ and Feature pyramid P$_{i}$ and then 
downsampling it. This process is iterative and terminated after I$_{5}$ is generated. 
This can be represented as
\begin{equation}
I_{l} = F_{l}([D_{l}(I_{l-1}), P_{l}])
\end{equation}
Where D$_{l}$ represents the downsampling function using strided 3x3 convolution.
On the other hand, the Bottom-up path augmentation of PANet uses element-wise
addition to combine layers. This is represented as
\begin{equation}
I_{l} = D_{l}(I_{l-1}) + P_{l}
\end{equation}
We can observe that both equations 7 and 8 are recursive. In equation 8 
(Bottom-up augmentation),
as the layers proceed, more features are added. However in our model (equation 7),
in addition to this we notice that the function
F$_{l}$ has more features to choose from due to concatenation with P$_{l}$ compared 
to bottom-up path augmentation method which doesn't use concatenation.
Features from all of the previous 
layers get reused similar DenseNet. Hence we are able to achieve better performance 
for the same computational cost of mask RCNN and for lower computational cost of 
bottom-up path augmentation.In our model, the combined Inception and Feature 
Pyramid layers form the layers
of the feature pyramid. Post hoc 3x3 convolutions are applied to these 
layers to reduce aliasing caused by upsampling. 

\subsection{Other Feature Pyramid Architectures}
\label{ssec:subhead}

In the model discussed previously, we have used concatenation to combine 
layers in the Bottom-up pathway and at the input to the Inception module.
Apart from the model (Model 1) discussed previously, we have experimented on 3 
other models to understand the effect of using concatenation to combine
layers. The first model (Model 2) uses concatenation to combine layers in the 
Bottom-up pathway and uses addition to combine layers for the Inception module 
input. That is, we use the bottom-most layer of the top-down pathway which 
is the element-wise sum of all other layers as an input to the Inception module.

The next model (Model 3), uses element-wise addition to combine layers in the bottom-up 
pathway. This method is similar to the layer combining process used in the Top-down
pathway except that we downsample the lower layer using strided 3x3 convolution
and then add it to the upper layer using element-wise addition. This model can be 
seen as bottom-up augmentation of PANet with our Inception module. The last model 
(Model 1*) is similar to the Model 1 except that it does not contain the 3x3 
post-hoc convolution which is used to reduce the aliasing effect of upsampling.

\begin{figure}[htb]
\begin{minipage}[c]{1\linewidth}
  \centering
  \centerline{\includegraphics[width=4.5cm]{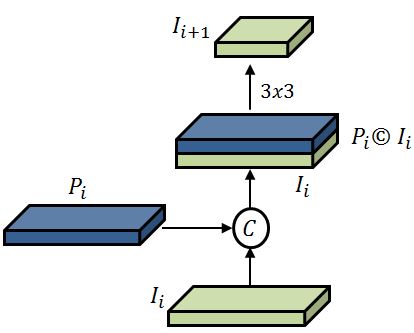}}
\end{minipage}
\caption{Detailed illustration of Bottom-up pathway.}
\label{fig:res}
\end{figure}
\begin{table*}[t]
\centering
\begin{tabular}{ |p{2.5cm}||p{0.8cm}|p{0.8cm}|p{0.8cm}|p{0.8cm}|p{0.8cm}|p{0.8cm}||p{0.8cm}|p{0.8cm}|p{0.8cm}|p{0.8cm}|p{0.8cm}|p{0.8cm}|}
 \hline
 \multicolumn{1}{|c||}{ } & \multicolumn{6}{c||}{Mask} & \multicolumn{6}{c|}{Bounding Box}\\
 \hline
 Model &AP&AP$_{50}$ &AP$_{75}$&AP$_{S}$&AP$_{M}$&AP$_{L}$&AP&AP$_{50}$&AP$_{75}$&AP$_{S}$&AP$_{M}$&AP$_{L}$\\
 \hline
 Ours:Model 1  & \textbf{34.8}  & 56.3  & \textbf{37.1}  & 15.4  & \textbf{37.5}  & 52.4 & \textbf{38.3}  & 59.7  & \textbf{41.7}  & \textbf{22.1}  & 41.2  & 50.9\\
 Ours:Model 2  & 34.7  & 56.4  & 36.9  & \textbf{15.5}  & 37.2  & 52.2 & \textbf{38.3}  & 59.6  & 41.6  & 21.9  & 41.3  & 50.9\\
 Ours:Model 3  & 34.5  & 56.0  & \textbf{37.1}  & 15.3  & 36.8  & \textbf{52.5} & 38.1  & 59.1  & \textbf{41.7}  & 21.5  & 40.9  & \textbf{51.0}\\
 Ours:Model 1* & 34.6  & \textbf{56.8}  & 36.7  & 15.0  & 37.3  & 52.1 & 38.1  & \textbf{60.0}  & 41.3  & 21.6  & 
 \textbf{41.4}  & 50.3\\
 \hline
 MRCNN + BPA   & 34.4  & 56.1  & 36.4  & 15.1  & 36.9  & 50.8 & 38.0  & 59.2  & 41.2  & 21.5  & 41.0  & 50.1\\
 FCIS ++ \cite{C14}      & 33.6  & 54.5  &  -    &  -    &  -    &  -   &  -    &  -    &  -    &  -    &  -    &  -  \\
 \hline
 Baseline:MRCNN     & 33.9  & 56.0  & 35.6  & 15.1  & 36.4  & 51.2 & 37.8  & 59.4  & 40.9  & 21.6  & 40.7  & 49.9\\
 \hline
\end{tabular}
\centering
\caption{Comparison of our models with Mask RCNN and Mask RCNN with bottom-up path augmentation(BPA) on COCO
         dataset. Model 1* is Model 1 without post hoc convolutions.}
\end{table*}


\begin{table}[t]
\centering
\begin{tabular}{ |p{2.5cm}||p{2.5cm}|p{2.5cm}|}
 \hline
 Model &MAC Computations in FPN &FPN Parameters\\
 \hline
 Ours:Model 1  & 56.7x10$^{9}$  & 3.5x10$^{6}$\\
 Ours:Model 2  & 58.2x10$^{9}$  & 3.4x10$^{6}$\\
 Ours:Model 3  & 70.2x10$^{9}$  & 4.5x10$^{6}$\\
 Ours:Model 1* & \textbf{9.6x10$^{9}$}  & \textbf{1.1x10$^{6}$}\\
 Mask RCNN     & 52.2x10$^{9}$  & 2.6x10$^{6}$\\
 MRCNN + BPA   & 63.9x10$^{9}$  & 4.4x10$^{6}$\\
 \hline
\end{tabular}
\centering
\caption{Comparison of number of Multiply-Accumulate computations assuming 1200x800 image size and parameters.}
\end{table}

\section{Experiments}
\label{sec:typestyle}

\subsection{Dataset and Metrics}
\label{ssec:subhead}

We have implemented all of the model using Caffe2 and Detectron\cite{C12} framework.
COCO dataset \cite{C11} is one of the most popular dataset for instance segmentation
and also one of the most challenging with each image containing
multiple instances with complex spatial layout. The dataset consists of 115k
labeled images for training, 5k images for validation, 20k images for 
test dev and 20k images for test-challenge. It has 80 classes with pixel-wise 
instance annotation. We have trained all the models on train-2017 subset and reported
results on val-2017 subset.

\subsection{Hyper-parameters}
\label{ssec:subhead}

We trained all of the models with an image batch size of 4. Based on the image 
batch size, we have
used a learning rate of 0.005 for 300k iterations, 0.0005 for the next 100k iterations
and 0.00005 for the last 50k iterations. The learning rate and number of iterations 
are based on \cite{C13}. All models are trained and tested with batch
normalized \cite{C10} ResNet50 as backbone.
For faster and more efficient training, we have initialized our ResNet backbone with 
pretrained weights from ImageNet 1k\cite{C18}.

\subsection{Experimental Results}
\label{ssec:subhead}

Our first model (Model 1) gave the best precision overall and is closely followed by
second and third models. The first model improves mask AP and bounding box AP\cite{C25} 
by 0.9 and 0.5
respectively over Mask RCNN, and by 0.4 and 0.3 over Mask RCNN with bottom-up path
augmentation. In figure 3, we can observe that the boundaries of the masks 
are more likely to bound to the edges of the objects since low-level features are 
used optimally for all sizes of object proposal in contrast to only small object proposals
in Mask RCNN. This performance boost can be observed in AP$_{L}$ column of table 1, 
where it is the highest increase in AP.

Model 2 gives a slightly lower performance than Model 1. This shows that the layer
combination method for the Inception module is not critical. However both Model 1
and Model 2 give better performance compared to Model 3. Model 3 is equivalent to 
the bottom-up augmentation of PANet with our Inception module added. This proves that 
concatenation indeed gives better performance compared to element-wise addition 
if used to combine layers in a feature pyramid network.

We can observe that adding a post hoc convolution to reduce aliasing boosts 
precision, especially of small objects, but comes at the cost of increased 
computational overhead.
The first model without post hoc convolution has better performance over Mask RCNN and 
Mask RCNN with BPA. Most importantly, it has fewer parameters and a very small computational 
overhead - much smaller than the original FPN itself!

\begin{figure}%
    \centering
    \subfloat[Mask RCNN]{{\includegraphics[width=3.7cm]{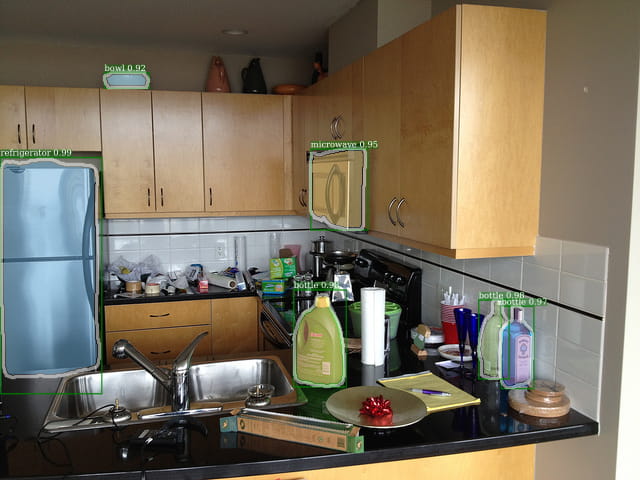} }}%
    \qquad
    \subfloat[Ours:Model 1]{{\includegraphics[width=3.7cm]{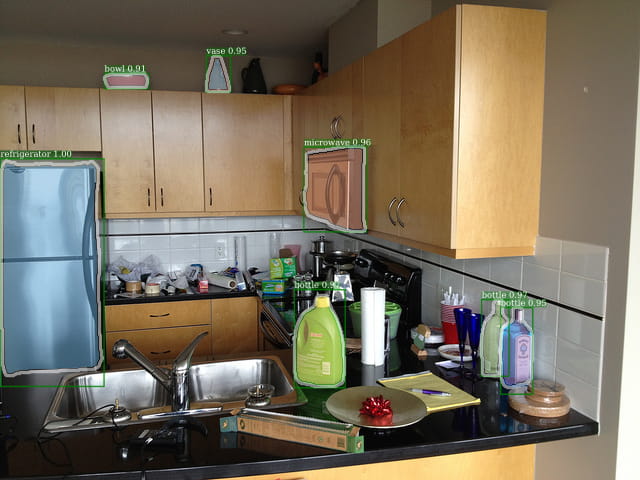} }}%
    \label{fig:example}%
\end{figure}
\begin{figure}%
    \centering
    \subfloat[Mask RCNN]{{\includegraphics[width=3.7cm]{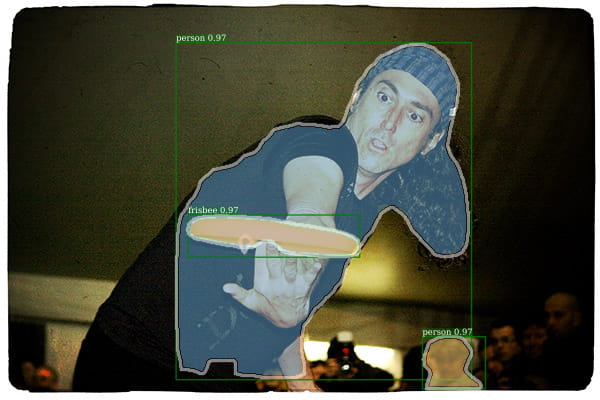} }}%
    \qquad
    \subfloat[Ours:Model 1]{{\includegraphics[width=3.7cm]{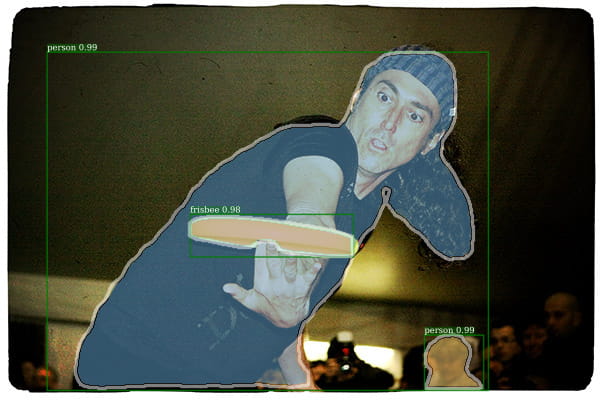} }}%
    \caption{Comparison of masks generated by Mask RCNN our model. Our model
        consistently gives better mask quality within the bounding boxes. Best viewed 
        electronically. Zoom in to see in more detail.}%
    \label{fig:example}%
\end{figure}

\section{Conclusion}
\label{sec:print}

In this paper, we propose a new framework to optimally infuse low-level features into 
higher pyramid levels and 
generate better quality masks. Our experiment results demonstrate that our 
model can improve the performance compared to Mask RCNN and Bottom-up path
Augmentation technique of PANet because our framework takes advantage of 
optimal combination of different levels of features of all layers of the 
feature pyramid. All of these improvements are done without any additional
computational cost.



\bibliographystyle{IEEEbib}
\bibliography{strings,refs}

\end{document}